\documentclass[11pt]{article}
\usepackage{eamt23}
\usepackage{times}
\usepackage{url}
\usepackage{latexsym}
\usepackage{xcolor}
\usepackage{float}
\usepackage[small,bf]{caption} 
\usepackage{booktabs}
\usepackage{float}
\usepackage{hyperref}
\usepackage{pgfplots}

\title{An Empirical Study of Leveraging Knowledge Distillation \\for Compressing Multilingual Neural Machine Translation  Models}

\author{
Varun Gumma\textsuperscript{\normalfont 1}, Raj Dabre\textsuperscript{\normalfont 2}, Pratyush Kumar\textsuperscript{\normalfont 3}\\
  Indian Institute of Technology, Madras\textsuperscript{\normalfont 1,3} \quad
  Microsoft\textsuperscript{\normalfont 3} 
  \quad AI4Bharat\textsuperscript{\normalfont 1,2,3}\\
  National Institute of Information and Communications Technology\textsuperscript{\normalfont 2}\\
  {\tt \textsuperscript{\normalfont 1}varun230999@gmail.com \quad \textsuperscript{\normalfont 2}raj.dabre@nict.go.jp} \\ {\tt \textsuperscript{\normalfont 3}pratykumar@microsft.com}
} 

\date{}

\begin{document}
\maketitle
\begin{abstract}
Knowledge distillation (KD) is a well-known method for compressing neural models. However, works focusing on distilling knowledge from large multilingual neural machine translation (MNMT) models into smaller ones are practically nonexistent, despite the popularity and superiority of MNMT. This paper bridges this gap by presenting an empirical investigation of knowledge distillation for compressing MNMT models. We take Indic to English translation as a case study and demonstrate that commonly used language-agnostic and language-aware KD approaches yield models that are $4$-$5\times$ smaller but also suffer from performance drops of up to $3.5$ BLEU. To mitigate this, we then experiment with design considerations such as shallower versus deeper models, heavy parameter sharing, multi-stage training, and adapters. We observe that deeper compact models tend to be as good as shallower non-compact ones, and that fine-tuning a distilled model on a High-Quality subset slightly boosts translation quality. Overall, we conclude that compressing MNMT models via KD is challenging, indicating immense scope for further research.
\end{abstract}
\section{Introduction}

Neural Machine Translation (NMT) \cite{DBLP:journals/corr/BahdanauCB14,NIPS2017_3f5ee243} is a state-of-the-art approach to machine translation that has gained significant attention in recent years. With the availability of large corpora and compute, Multilingual NMT (MNMT) \cite{zhang-etal-2019-bridging,firat-etal-2016-multi,aharoni-etal-2019-massively} has gained popularity since it enables a single model to translate between multiple languages. Large MNMT models trained on substantial data have shown higher levels of performance. However, these models are impractical for deployment on a commercial or production scale due to their size, which contains millions, if not billions, of parameters. Therefore, they need to be compressed into smaller models for efficient and convenient usage. 

\begin{figure}[t]
\centering
\includegraphics[width=0.5\textwidth]{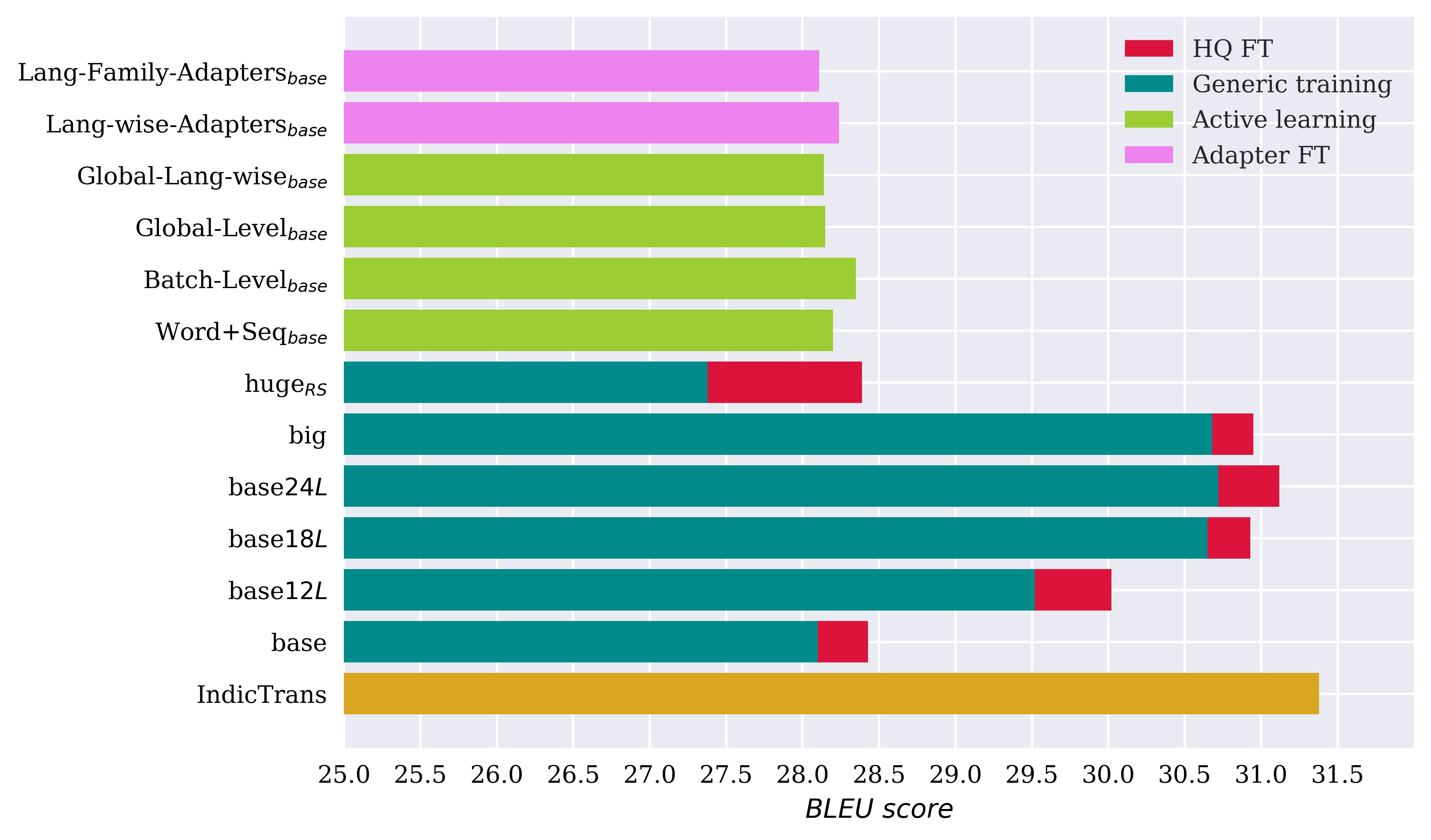}
\caption{A comparison of the major distillation techniques and models we experimented with. Note that the \textcolor{red}{red} increments in the bar plots denote the improvements due to HQ fine-tuning for those models.}
\end{figure}

In practice, models are compressed via two methods: Firstly, by stripping unnecessary and redundant parameters from the existing model \cite{10.1145/1150402.1150464}, and secondly, by transferring knowledge from the larger ``teacher" model to a smaller ``student" model using distillation \cite{https://doi.org/10.48550/arxiv.1503.02531}. This study focuses on the latter, as the former can be done post-hoc \cite{diddee-etal-2022-brittle}. Although existing literature mainly discusses bilingual-to-multilingual or bilingual-to-bilingual distillation, to the best of our knowledge, there is no work in end-to-end multilingual-to-multilingual knowledge distillation for compression in a setting with a mix of low, medium, and high resource languages. Therefore, we aim to distill a large MNMT model into a smaller one taking Indic to English language translation as a case study and perform an empirical investigation of prominent techniques such as language agnostic and language-wise word-level and sequence-level distillation. We also look into architectural variations, multi-stage training, and High-Quality data filtering to improve our performance.

\noindent Our contributions can be summarized as follows: \\
\noindent \textbf{1.} We investigate the effect of existing distillation techniques for compressing MNMT models and find that all of them produce comparable results, indicating that the simplest methods are sufficient. \\
\noindent \textbf{2.} We explore the outcome of language-specific architectures such as Adapters and Language-Queues and conclude that they failed to sufficiently specialize the models for significant gains. \\
\noindent \textbf{3.} We analyze the performance gains due to multi-stage training and find that High-Quality fine-tuning boosts performance in a noisy scenario. \\
\noindent \textbf{4.} We analyze the trade-off between width and height for Transformers \cite{NIPS2017_3f5ee243} and determine that thinner but deeper models comprise fewer parameters but perform comparably to wider but shallower models.
\section{Related works}

This paper focuses on Knowledge Distillation (KD) for compressing Multilingual Neural Machine Translation (MNMT) models. 

\noindent \textbf{Multilingual Neural Machine Translation} \cite{zhang-etal-2019-bridging,firat-etal-2016-multi,aharoni-etal-2019-massively} is the favored approach for developing machine translation systems that can handle multiple languages. MNMT systems incorporate language-specific information through the use of shared encoder and decoder architecture and language-specific embeddings. MNMT systems often require less training data than separate bilingual models for each language, making it an attractive area of research. A detailed analysis of MNMT can be found in the survey paper by \cite{dabre-etal-2020-multilingual}. \\
\noindent \textbf{Model compression}, which involves pruning or reparameterizing large models to reduce their sizes, has been explored in previous studies \cite{10.1145/1150402.1150464,wang-etal-2020-structured,behnke-heafield-2020-losing,behnke-etal-2021-efficient}. Orthogonally, compression can be achieved by heavy parameter sharing, especially across layers \cite{Dabre_Fujita_2019}. \cite{dabre-etal-2022-indicbart} have investigated this in their IndicBART work, demonstrating that a significant parameter reduction leads to decreased performance, but knowledge distillation can help overcome this gap. We also explore this parameter sharing across layers, noting that we focus on compressing larger models in higher resource settings. \\
\noindent \textbf{Knowledge Distillation} \cite{https://doi.org/10.48550/arxiv.1503.02531,kim-rush-2016-sequence} is yet another orthogonal approach for model compression, to extract essential information from a larger model and transfer it to a smaller model while minimizing the drop in performance. \cite{dabre-fujita-2020-combining} present an approach leveraging Sequence-Level Distillation \cite{kim-rush-2016-sequence} with Transfer Learning for efficiently training NMT models in a highly low-resource scenario. However, their setup focused on relatively minor data scales, whereas we mainly operate in a medium to high resource scenario with multilingualism. \cite{10.1145/3546067} propose a multilingual distillation technique but use multiple multilingual strong teacher models of similar languages, similar to the method of \cite{tan2018multilingual} where they employ bilingual teacher models to distill into a single multilingual student. Our work differs from both in two aspects: (a) we do not use multiple bilingual/multilingual models as teachers, but instead focus on distilling one single robust multilingual model into another multilingual model end-to-end (b) we aim to compress where they do not. We do not use their techniques because our preliminary investigations showed that our teacher model was better than individual bilingual or multilingual models of similar languages. \\ \\
To the best of our knowledge, previous research on distillation has focused on distilling bilingual networks or training an equally sized student model from multiple strong bilingual/multilingual teacher models. Therefore, we believe our work is a first-of-its-kind introductory investigation in the domain of end-to-end distillation of MNMT models for compression. 
\section{Methodology}
This section describes the KD approaches and design considerations we focused on in this paper.

\begin{figure*}[t]
\centering
\includegraphics[width=0.85\textwidth]{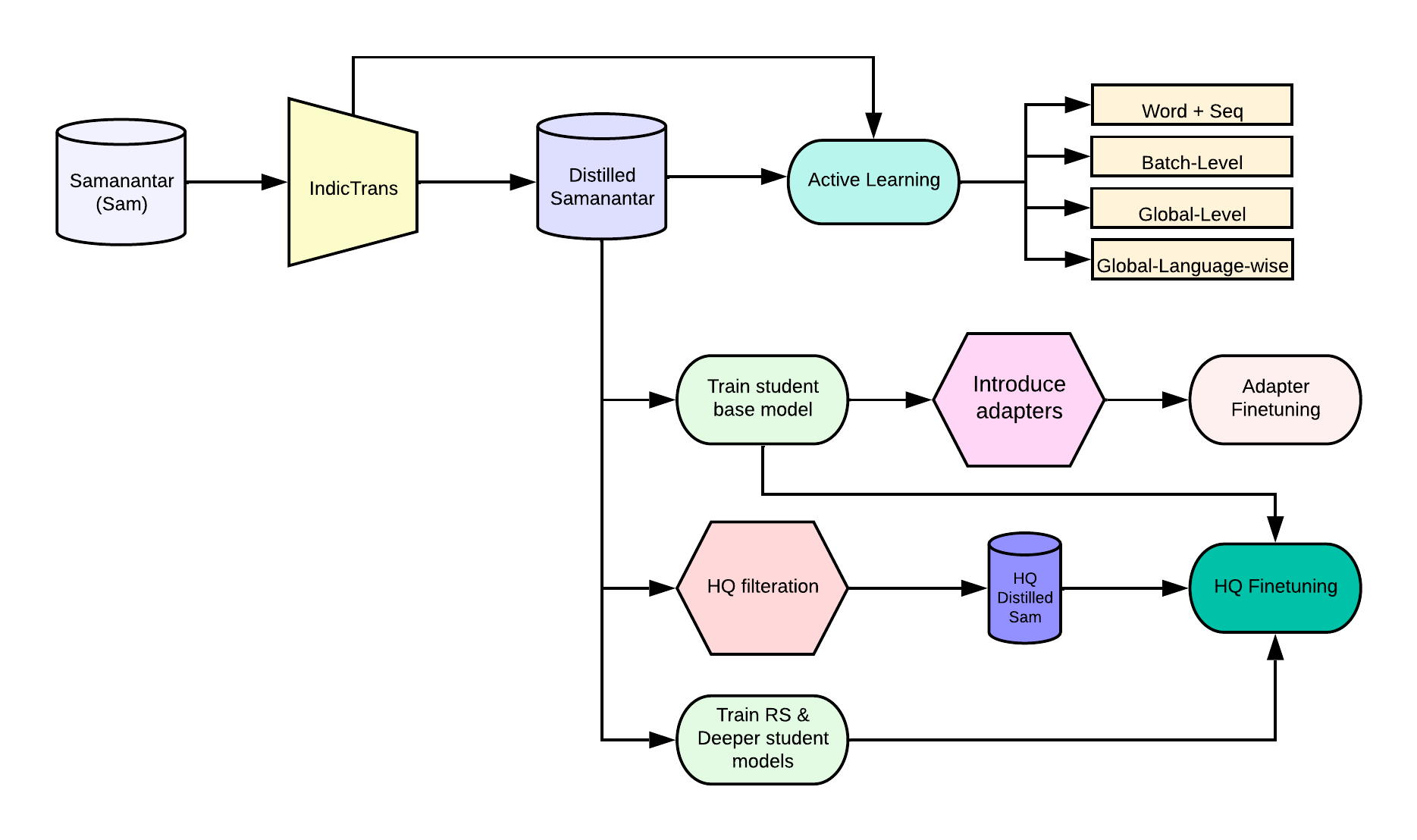}
\caption{A flow chart depicting our set of experiments}
\end{figure*}

\subsection{KD Approaches}
We describe the fundamental language-agnostic KD approaches, such as word and Sequence-Level KD and a language-aware KD approach using queues.

\noindent \textbf{Word-Level Distillation (WLD)}: 
Following \cite{https://doi.org/10.48550/arxiv.1503.02531}, \cite{kim-rush-2016-sequence} proposed Word-Level Distillation, which aims to minimize the KL-Divergence/Cross-Entropy between the student and teacher models at each time-step. However, we did not test this method because \cite{kim-rush-2016-sequence} showed that it is not a good approximation of the sequential learning task, as it focuses on the current timestep only and not on the entire sequence.

\noindent \textbf{Sequence-Level Distillation (SLD)}: \cite{kim-rush-2016-sequence} argued that the student model should capture the Sequence-Level distribution of the teacher model rather than the individual word-level distribution at each timestep. Therefore, they proposed that capturing the best beam search output of the teacher, which can approximate the distribution, can be used as hard pseudo-labels for the student. These hard pseudo-labels are called the \textit{distilled} targets. We extensively used this Sequence-Level Distillation technique to train all our student models because it is easy to implement and has been proven to give better results than regular word-level distribution.

\noindent \textbf{Word + Sequence-Level Distillation (W+S LD)}: \cite{kim-rush-2016-sequence} further proposed that Word-Level Distillation can be carried out in congruence with Sequence-Level Distillation to aid the student model in capturing both the word-level distribution at each timestep and the overall Sequence-Level distribution. This allows the student model to mimic the generalization of the teacher better. Hence, we applied this technique to determine if there were any improvements in performance over vanilla Sequence-Level Distillation.

\noindent \textbf{Selective Distillation}: \cite{wang-etal-2021-selective} showed that some samples are ``hard" to distill and require additional distillation signals to train, while others are ``easy" and do not. Therefore, they proposed the idea of identifying ``hard" samples from a batch and applying a word-level distillation loss specifically to them. They further extended the Batch-Level selection to Global-Level selection, where they select ``hard" samples from a large queue comparable in size to the entire dataset to better approximate the negative log-likelihood loss distribution used to identify ``hard" samples. Since we operate with a mix of low, medium, and high-resource languages, we chose to investigate both their \textbf{Batch-Level (BL)} and \textbf{Global-Level (GL)} selection strategies to promote low-resource languages, which might be challenging to distill due to their scarcity during training.

\noindent \textbf{Global-Language-wise Distillation (GLwD)}: The selection strategy proposed by \cite{wang-etal-2021-selective} at the global level is designed for bilingual settings. However, in multilingual settings with mixtures of languages with varying levels of abundance, a single global queue may not be suitable because it may become populated with samples mainly from high-resource languages. As a result, the selection algorithm may be biased toward resource-rich languages. Therefore, we propose a novel modification to this technique involving a language-wise selection strategy. Specifically, we propose to push samples from each language into their respective global queues, remove the oldest samples to maintain the queue size, and apply an additional distillation loss to the ``harder" samples from each queue, similar to the Global-Level selection.

\subsection{Design Considerations}
\label{sec:design_considerations}
Apart from the core distillation approaches above, we also explore the impact of several architectural and training pipeline design considerations. In particular, we focus on the impact of variable depth, extreme parameter-sharing, dataset filtering and multi-stage training, and language-specific distillation via adapters.

\noindent \textbf{Width vs. Height:} Based on the findings of \cite{tay2022scale}, we opted to analyze thinner but deeper models, as we found these models to have fewer parameters than wider but shallower models.

\noindent \textbf{Recurrent-Stacking:} We also train models on the \textit{distilled} data with recurrently stacked layers, following the idea of \cite{Dabre_Fujita_2019} in which layer parameters are tied across layers. This limited the number of parameters to $207$M but gave the effect of a model with multiple layers.

\noindent \textbf{Multi-stage Training with High-Quality Data:} We observed that the distilled data contained a few noisy samples that hindered training. To address this issue, we experimented with a multi-stage training setup. First, we trained a smaller model on the complete dataset, and then we fine-tuned it on the High-Quality data filtered from the complete dataset. We filtered the data based on the LaBSE\footnote{\url{https://huggingface.co/sentence-transformers/LaBSE}} \cite{feng-etal-2022-language} cosine similarity scores, selecting only those translation pairs whose similarity score was greater than $\mu_L + k\sigma_L$ for each language, where $u_L$ and $\sigma_L$ denote the mean and standard deviation of the translation scores for language $L$. We empirically chose $k$ to limit the High-Quality data size to approximately $20\%$ of the total, with a uniform sampling of data from each language. 

\noindent \textbf{Adapters:} Adapters are small feed-forward modules introduced in pre-trained models and fine-tuned on a downstream task while freezing the trained model's parameters \cite{pmlr-v97-houlsby19a,bapna-firat-2019-simple}. They add only a tiny fraction of parameters to the model but provide additional parameterization for the model to adapt to additional languages/domains independently without requiring complete fine-tuning. Adapters are particularly useful for distillation, as they should help recover any loss in performance due to compression via fewer additional parameters. Furthermore, they should help the model adjust to various languages' specifics during translation. To investigate the effects of language similarity and cross-lingual inference on distillation, we have experimented with fine-tuning distilled models with adapters for individual languages and language families \cite{https://doi.org/10.48550/arxiv.2209.15236}.

\section{Experiments}
We now focus on Indic-to-English translation as a case study and describe experiments we conducted to compress IndicTrans, a $474$M parameter model.

\subsection{Datasets}
We use or create the following datasets: 

\noindent \textbf{Original data}: We use Samanantar \cite{ramesh-etal-2022-samanantar} as the original (undistilled) dataset, the statistics for which are in Table-\ref{table:sam-pairs} in the column \#Pairs. This dataset was used to train IndicTrans, our teacher model, and we use it for generating the \textit{distilled} data and conducting comparative studies.

\begin{table}[h]
\centering
\small
\begin{tabular}{@{}lccc@{}}
\toprule
\textbf{Lang} & \textbf{ISO code} & \textbf{\#Pairs} & \textbf{\#HQ Pairs} \\ \midrule
Assamese      & as                & 0.1              & 0.02                \\
Odia          & or                & 1.0                & 0.2                 \\ \midrule
Punjabi       & pa                & 3.0                & 0.6                 \\
Gujarati      & gu                & 3.1              & 0.6                 \\
Marathi       & mr                & 3.6              & 0.8                 \\
Kannada       & kn                & 4.1              & 0.9                 \\
Telugu        & te                & 4.9              & 1.1                 \\
Tamil         & ta                & 5.3              & 1.0                   \\
Malayalam     & ml                & 5.9              & 1.3                 \\ \midrule
Bengali       & bn                & 8.6              & 1.7                 \\
Hindi         & hi                & 10.1             & 2.0                   \\ \midrule
Total         & -                 & 49.8             & 10.3                \\ \bottomrule
\end{tabular}
\caption{The number of original (\#pairs) sentence pairs per language (in millions) in the \textit{distilled} (and original). \#HQ-Pairs indicates High-Quality distilled pairs. The languages are categorized into low, medium, and high-resource groups.}
\label{table:sam-pairs}
\end{table}

\noindent \textbf{Distilled data}: The \textit{distilled} data used for training student models was generated by performing beam search (with a beam size of $5$) over Samanantar in the Indic-En direction with IndicTrans., i.e., using the Sequence-Level distillation technique of \cite{kim-rush-2016-sequence}. The best beam output was then utilized as the hard pseudo-labels for training smaller models. Following Section~\ref{sec:design_considerations}, we filter this data to obtain a smaller, higher quality version, the statistics for which are in the column \#HQ-Pairs in Table-\ref{table:sam-pairs}.

\noindent \textbf{Evaluation data}: We use Flores101 \cite{goyal-etal-2022-flores} for evaluation, where the dev set ($997$ pairs per language) is used for validation and the test set ($1012$ pairs) for testing.

\subsection{Pre-Processing and Vocabulary}
We follow \cite{ramesh-etal-2022-samanantar} and transliterate all the Indic source sentences into Devanagari using the Indic-NLP-Library\footnote{\url{https://github.com/anoopkunchukuttan/indic_nlp_library}} before training, to take advantage of the script-similarity between various Indian languages. 
The dev-test set is likewise transliterated, and language tags are added before evaluation. For consistency, we use the same vocabulary as IndicTrans, which contains $32$K subwords for all $11$ Indic languages and separate $32$K subwords for English.

\subsection{Evaluation Metrics}
We use BLEU \cite{papineni-etal-2002-bleu} as the primary evaluation metric. We also report Chrf++ scores \cite{popovic-2017-chrf} in the Appendix.

\subsection{Training setup}
We train our models using \textit{fairseq}\footnote{\url{https://github.com/VarunGumma/fairseq}} \cite{ott-etal-2019-fairseq}. We obtained the implementation for KD from LeslieOverfitting\footnote{\url{https://github.com/LeslieOverfitting/selective_distillation}}. 
The Transformer architecture \cite{NIPS2017_3f5ee243} is used throughout our experiments. The hyperparameters used for training are presented in Appendix-\ref{appendix-a} Table-\ref{table:training-hyperparams}.

Unlike IndicTrans, we use GELU activation \cite{https://doi.org/10.48550/arxiv.1606.08415} instead of ReLU activation. Additionally, pre-normalization is applied to all modules, and layer normalization \cite{https://doi.org/10.48550/arxiv.1607.06450} is applied to the embedding. These modifications led to more stable training. Where early stopping for IndicTrans was done using loss on the development set, we used BLEU score.

\subsection{Model Configurations}

We trained models with various configurations (as listed in Table-\ref{table:model-specs}). The smallest model is ``base'', the same as Transformer-base in \cite{NIPS2017_3f5ee243}. The largest is ``huge" which is the same size as IndicTrans, and ``huge$_{RS}$" is its equivalent where all layers have the same parameters.

\begin{table}[h]
\centering 
\small
\begin{tabular}{@{}lccccc@{}}
\toprule
\textbf{Model} & \textbf{P} & \textbf{d$_{M}$} & \textbf{d$_{FF}$} & \textbf{L} & \textbf{H} \\ \midrule
base           & 95.4                  & 512               & 2048              & 6                  & 8                 \\
base$12L$      & 139.5            & 512               & 2048              & 12                 & 8                 \\
base$18L$      & 183.7              & 512               & 2048              & 18                 & 8                 \\
base$24L$      & 227.8             & 512               & 2048              & 24                 & 8                 \\
big      & 278.9                 & 1024              & 4096              & 6                  & 16                \\
huge$_{RS}$  & 207.3                & 1536             & 4096              & 1                  & 16                \\
huge       & 474.9          & 1536              & 4096              & 6                  & 16                \\ \bottomrule
\end{tabular}
\caption{The table presents the architectural description of various Transformer models that were tested. Here, the columns represent the number of parameters (P) in millions, the dimension of the model (d$_{M}$), the dimension of the feed-forward network (d$_{FF}$), the number of layers (L) and the number of attention heads (H). It is worth noting that the huge$_{RS}$ model contains only one unique layer, but it is recurrently stacked $6$ times. This means the other $5$ layers in the encoder/decoder are simply references to the original layer.}
\label{table:model-specs}
\end{table}
\section{Results}
This section presents the results of applying Knowledge Distillation (KD) approaches to compress the IndicTrans Indic-to-English teacher model. 

\subsection{Main Results}
Table-\ref{table:distillation-bleu-results} compares various distillation approaches using a student model with the \textit{base} configuration. As compared to a base model trained on the original data, which is around $3.6$ BLEU below the IndicTrans model, we can observe improvements for both low and high-resource languages through the use of conventional distillation methods. The simplest among these, Sequence-Level distillation (SLD), shows an improvement of $0.3$ BLEU on average compared to its undistilled equivalent. Significantly, low-resource languages such as Assamese and Odia and a few medium-resource languages like Kannada benefit the most. In contrast, resource-rich languages like Hindi and Bengali have comparable or a slight drop in performance. The Batch-Level selection approach (BL) was the best among all distillation approaches and showed the best results for $6$ out of $11$ languages. On the other hand, Global-Level selection (GL) did not perform as well, indicating that adaptation is best done per batch since Global-Level selection may update similar examples whereas Batch-Level adaptation would choose diverse examples. Further, we observed that the queue size should be meticulously tuned in case of a mix of languages. 

To our surprise, active distillation (W+S LD) failed to significantly improve despite leveraging distilled data and the parent model's soft labels. Also, or adaptation of Global-Level selection to Global-Language-wise Distillation (GLwD) resulted in only minor variations when compared to the base model that was trained using regular Sequence-Level distillation and Global-Level distillation. Interested readers can check Chrf++ scores in Appendix-\ref{appendix-b}, Table-\ref{table:distillation-chrf-results}, and observe that they follow the same trend. 

No matter the approach, however, the distilled model consistently underperforms the teacher, indicating the high difficulty of distilling MNMT models. Indeed, where the \textit{base} model trained without distilled data was behind by $3.6$ BLEU, the best-distilled model is behind by $3.1$ BLEU on average. Going forward, for the ease of rapidly conducting large-scale experiments, we only report and discuss the results of remaining models trained using Sequence-Level distillation, i.e., by directly training them on the \textit{distilled} dataset.

\begin{table}[t]
\centering 
\small
\setlength{\tabcolsep}{2pt}
\begin{tabular}{@{}ccc|ccccc@{}}
\toprule
\textbf{Lang} & \textbf{OG\_base} & \textbf{IT} & \textbf{SLD} & \textbf{W+S LD} & \textbf{BL} & \textbf{GL} & \textbf{GLwD} \\ \midrule
as            & 18.4        & 23.3        & 19.7         & 19.8            & \textbf{20.5}        & 20.3        & \textbf{20.5}          \\
bn            & 28.9        & 31.8        & 28.8         & 28.9            & \textbf{29.1}        & 28.3        & 28.7          \\
gu            & 30.6        & 34.1        & 30.6         & 31.5            & \textbf{31.7}        & 31.3        & 30.9          \\
hi            & 34.3        & 37.5        & 34.1         & 34.2            & \textbf{34.7}        & 34.4        & 34.6          \\
kn            & 25.2        & 28.7        & \textbf{26.1}         & 25.8            & 25.9        & 26.0          & 25.8          \\
ml            & 27.7        & 31.4        & \textbf{28.2}         & 27.9            & \textbf{28.2}        & 27.6        & 28.0            \\
mr            & 27.4        & 31.0          & \textbf{28.1}         & 28.0              & 27.8        & 27.5        & 27.8          \\
or            & 26.3        & 29.8        & 26.8         & 27.0              & 27.0          & \textbf{27.1}        & 26.5          \\
pa            & 31.0          & 35.8        & 31.2         & \textbf{31.4}            & 31.3        & \textbf{31.4}        & 31.1          \\
ta            & 25.3        & 28.4        & 25.1         & 25.1            & \textbf{25.4}        & 25.2        & 25.2          \\
te            & 30.4        & 33.4        & 30.4         & \textbf{30.6}            & 30.2        & \textbf{30.6}        & 30.4          \\ \midrule
Avg           & 27.8       & 31.4       & 28.1         & 28.2            & 28.3       & 28.2       & 28.1         \\ \bottomrule
\end{tabular}
\caption{BLEU scores of base model \textit{distilled} with various distillation techniques. Note that the scores of the \textit{base} model trained on the Original Samanantar data (OG\_base) and IndicTrans (IT; \textit{huge}) in the first and second columns are for reference. The best scores of distilled models are bolded.}
\label{table:distillation-bleu-results}
\end{table}

\subsection{Analyses and Further Investigation}
We now investigate factors that influence distillation. We analyze the quality of the distillation data, the impact of different model architectures, and multi-stage training using High-Quality data for further training models or with adapters without High-Quality data. These experiments can help us ascertain whether the poor performance of distilled models can be remedied.

\noindent \textbf{Distilled Dataset Analysis:} LaBSE cosine-similarity scores were used to assess the quality of translation pairs in the \textit{distilled} data. The \textit{distilled} dataset was significantly better, as evidenced by higher mean and lower standard deviation of the LaBSE scores, as shown in Table-\ref{table:og-distilled-labse}.

\begin{table}[h]
\centering 
\small
\begin{tabular}{@{}c|cc|cc@{}}
\toprule
\textbf{}           & \multicolumn{2}{c}{\textbf{OG}}   & \multicolumn{2}{|c}{\textbf{Distilled}} \\ \midrule
\textbf{Lang\_pair} & \textbf{mean} & \textbf{std\_dev} & \textbf{mean}    & \textbf{std\_dev}   \\ \midrule
en-as               & 0.6460         & 0.2773           & 0.7850            & 0.1297              \\
en-bn               & 0.7974        & 0.1286            & 0.8446           & 0.0726              \\
en-gu               & 0.8007        & 0.1515            & 0.8487           & 0.0699              \\
en-hi               & 0.7988        & 0.1159            & 0.8524           & 0.0737              \\
en-kn               & 0.8129        & 0.1240            & 0.8469           & 0.0680               \\
en-ml               & 0.8018        & 0.1310            & 0.8432           & 0.0743              \\
en-mr               & 0.7886        & 0.1471            & 0.8472           & 0.0672              \\
en-or               & 0.8283        & 0.0877            & 0.8474           & 0.0666              \\
en-pa               & 0.7958        & 0.1383            & 0.8579           & 0.0726              \\
en-ta               & 0.7762        & 0.1691            & 0.8415           & 0.0771              \\
en-te               & 0.8152        & 0.1089            & 0.8448           & 0.0685              \\ \bottomrule
\end{tabular}
\caption{LaBSE cosine similarity scores between translation pairs of Original and Distilled data}
\label{table:og-distilled-labse}
\end{table}

\noindent \textbf{Impact of Deeper vs. Shallower Models on Performance and Inference Time:} Table-\ref{table:width-vs-depth-perf} shows that thinner but deeper networks perform comparably with the wider but shallower models while having fewer parameters. However, Table-\ref{table:width-vs-depth-inference-time} also highlights that the deeper models often suffer from longer latency during inference due to the numerous sequential transformations to the input in both the encoder and decoder. Furthermore, we observed diminishing returns in performance as we increased the number of layers.

\noindent \textbf{Impact of extreme parameter sharing:} From Table-\ref{table:width-vs-depth-perf} we can see that recurrent stacking (\textit{huge$_{RS}$}) is not particularly impactful. Note that the key difference between the \textit{huge} and \textit{huge$_{RS}$} models is that the latter has shared layer parameters. \cite{dabre-etal-2022-indicbart} showed that recurrent stacking models, when trained with distillation data, can reach the performance of the parent model (\textit{huge}), but this does not appear to be the case in our setting. Note that, in our case, our training data is much larger than \cite{dabre-etal-2022-indicbart}, indicating that recurrent stacking models might not be suitable here.
Next, the inference time for \textit{huge$_{RS}$} is almost the same as its \textit{huge} counterpart because the input is still transformed the same number of times, but just using the same layer. Comparing with the deeper base models (\textit{base12L}, \textit{base18L}, \textit{base24L}), increasing the width of models increases parameters but results in only a slight increase in inference times, unlike increasing the depth of the network.

\begin{table}[h]
\centering 
\small
\begin{tabular}{@{}ccccc@{}}
\toprule
\textbf{Lang} & \textbf{huge$_{RS}$} & \textbf{base$12L$} & \textbf{base$18L$} & \textbf{base$24L$} \\ \midrule
as                 & 19.2                   & 21.6          & \textbf{23.3}            & 22.9           \\
bn                 & 27.9                   & 29.8          & 30.9            & \textbf{31.1}           \\
gu                 & 30.4                   & 32.5          & 33.9            & \textbf{33.9}           \\
hi                 & 34.1                   & 36.0            & \textbf{36.6}            & 36.2           \\
kn                 & 25.4                   & 27.0            & \textbf{28.3}            & 28.0             \\
ml                 & 26.7                   & 29.3          & 29.8            & \textbf{30.5}           \\
mr                 & 26.7                   & 29.5          & 30.4            & \textbf{30.6}           \\
or                 & 25.4                   & 28.3          & 29.5            & \textbf{29.6}           \\
pa                 & 31.2                   & 33.0          & 34.0              & \textbf{34.2}           \\
ta                 & 24.6                   & 26.3          & 27.4            & \textbf{27.9}           \\
te                 & 29.6                   & 31.4          & \textbf{33.0}              & \textbf{33.0}             \\ \midrule
Avg                & 27.4                  & 29.5         & 30.6           & 30.7          \\ \bottomrule
\end{tabular}
\caption{Performance of models with varying depth}
\label{table:width-vs-depth-perf}
\end{table}

\begin{table}[t]
\centering
\small
\begingroup
\setlength{\tabcolsep}{1.25pt} 
\begin{tabular}{@{}cccccccc@{}}
\toprule
\textbf{Lang} & \textbf{base} & \textbf{base$12L$} & \textbf{base$18L$} & \textbf{base$24L$} & \textbf{big} & \textbf{huge$_{RS}$} & \textbf{huge} \\ \midrule
as            & 8.3           & 15.7             & 19.4             & 25.9             & 9.4          & 9.9               & 15.8          \\
bn            & 7.8           & 13.1             & 18.8             & 23.7             & 8.6          & 9.2               & 8.8           \\
gu            & 8.9           & 13.4             & 18.2             & 25.6             & 8.4          & 9.1               & 9.9           \\
hi            & 8.8           & 13.0             & 18.4             & 24.2             & 10.7         & 9.3               & 8.7           \\
kn            & 12.4          & 13.1             & 18.5             & 23.6             & 9.8          & 9.1               & 9.0           \\
ml            & 8.7           & 13.8             & 20.7             & 26.2             & 9.7          & 9.0               & 9.0           \\
mr            & 9.1           & 12.9             & 18.0             & 24.4             & 8.9          & 9.2               & 8.9           \\
or            & 9.2           & 13.7             & 20.9             & 24.3             & 9.3          & 9.4               & 9.0           \\
pa            & 8.9           & 13.7             & 19.3             & 24.7             & 8.9          & 9.2               & 9.0           \\
ta            & 8.4           & 13.4             & 20.3             & 23.8             & 8.7          & 9.8               & 9.4           \\
te            & 8.0           & 13.0             & 20.1             & 26.1             & 8.6          & 10.2              & 9.0           \\ \midrule
Avg           & 9.0           & 13.5             & 19.4             & 24.8             & 9.2          & 9.4               & 9.7           \\ \bottomrule
\end{tabular}
\endgroup
\caption{Inference time per language (in seconds) with a batch size of $64$ on the Flores101 test set ($1012$ sentences per language). As seen from the above table, \textit{base24L} has the highest latency due to the highest number of layers in the encoder and decoder.}
\label{table:width-vs-depth-inference-time}
\end{table}

\noindent \textbf{Multi-stage training:} The rationale behind High-Quality data fine-tuning is that it enables the model to relearn the richer set of examples and disregard the previously noisy examples, which hurt the performance.  We observed that the performance of the model improves with fine-tuning\footnote{For optimal fine-tuning, it is recommended to use a lower learning rate ($3e$-$5$) and a smaller batch size ($24$K).} an existing distilled model with HQ data (see Table-\ref{table:HQ-FT}). The maximum improvement was observed for the Recurrent Stacked model, which showed the weakest performance thus far, given its size. Note the improvement of the \textit{base} model from $28.1$ (SLD in Table~\ref{table:distillation-bleu-results}) to $28.4$, by $0.3$ BLEU. The previous gap between the parent (IndicTrans; \textit{huge}) and \textit{base} model was $3.3$, and it has now come down to $3.0$, indicating that the gap can be overcome, but that multilingual model compression is still very challenging.  

\begin{figure}[h]
\centering 
\includegraphics[width=0.5\textwidth]{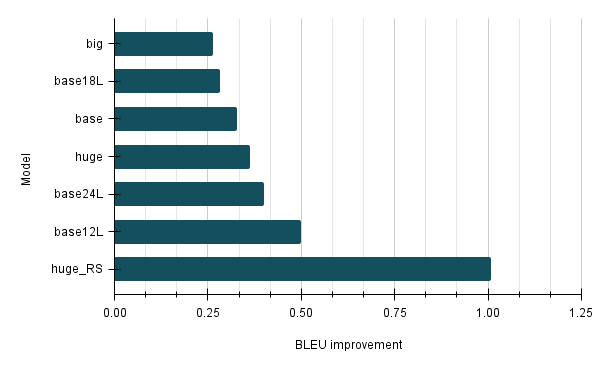}
\includegraphics[width=0.5\textwidth]{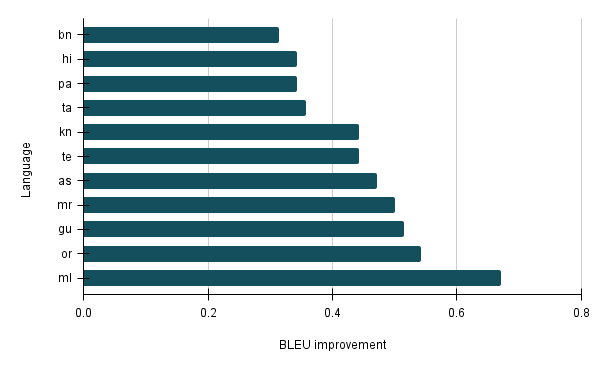}
\caption{\emph{Top:} Comparative bar plot of improvements due to HQ fine-tuning averaged over various languages vs. Model\\\emph{Bottom:} Comparative bar plot of improvements due to HQ fine-tuning averaged over various models vs. Language}
\label{figure:HQ-FT}
\end{figure}

The increments resulting from High-Quality fine-tuning were averaged across multiple models and languages, and the findings are presented in Figure-\ref{figure:HQ-FT}. It is observed in Figure-\ref{figure:HQ-FT} that multi-stage training had the least effect on high-resource languages such as Bengali and Hindi since the model well learned these languages due to the ample amount of training data available. Conversely, low-resource languages, such as Odia and Assamese, benefited from multi-stage training. Our analysis showed that Malayalam experienced the most significant improvement with HQ fine-tuning.

\begin{table}[h]
\centering
\small
\begingroup
\setlength{\tabcolsep}{1.25pt} 
\begin{tabular}{@{}ccccccc@{}}
\toprule
\textbf{Lang} & \textbf{base} & \textbf{base$12L$} & \textbf{base$18L$} & \textbf{base$24L$} & \textbf{big} & \textbf{huge$_{RS}$} \\ \midrule
as            & 0.6           & 0.7                & 0.3                & 0.3                & -0.1         & 1.2                  \\
bn            & 0.2           & 0.5                & 0.3                & 0.5                & -0.1         & 0.7                  \\
gu            & 0.6           & 0.6                & 0.1                & 0.2                & 0.4          & 1.1                  \\
hi            & 0.2           & 0.1                & 0.2                & 0.4                & 0.0            & 1.0                    \\
kn            & 0.3           & 0.6                & 0.2                & 0.5                & 0.2          & 0.8                  \\
ml            & 0.5           & 0.6                & 0.8                & 0.6                & 0.4          & 1.4                  \\
mr            & 0.0             & 0.5                & 0.4                & 0.3                & 0.7          & 1.2                  \\
or            & 0.5           & 0.6                & -0.2               & 0.3                & 0.9          & 1.3                  \\
pa            & 0.3           & 0.3                & 0.4                & 0.6                & -0.2         & 1.0                    \\
ta            & 0.2           & 0.6                & 0.1                & 0.2                & 0.3          & 0.8                  \\
te            & 0.2           & 0.4                & 0.5                & 0.5                & 0.4          & 0.6                  \\ \midrule
Avg           & 0.3          & 0.5                & 0.3               & 0.4                & 0.3         & 1.0                 \\ \bottomrule
\end{tabular}
\endgroup
\caption{Multistage training improvements. Once again, all these models were trained and fine-tuned on the \textit{distilled} dataset. The absolute scores, i.e., score of model trained on the distilled data + the increment by fine-tuning on HQ-distilled data is available in Table-\ref{hqft-abs-bleu} of Appendix-\ref{appendix-b}}
\label{table:HQ-FT}
\end{table}

\noindent \textbf{Adapters:} Adapters were introduced on top of the distilled \textit{base} model for each language and prominent language families, such as Eastern Indo-Aryan (Assamese-Bengali-Odiya), Western Indo-Aryan (Hindi-Gujarati-Punjabi-Marathi), and Dravidian (Kannada-Malayalam-Tamil-Telugu). Notably, these adapters were again fine-tuned on the unfiltered \textit{distilled} dataset. As presented in Table-\ref{table:adapter-FT}, the outcomes revealed that the language-wise and language-family adapters exhibited minimal or no improvement in the given setting. This lack of improvement could be attributed to the inadequacy of the added parameters in learning new representations from languages to enhance performance. Language-wise adapters outperformed language-family adapters since high-resource languages dominate the low-resource ones when building language families. In other words, when working with adapters, their limited capacity can only handle limited data. Although we do not show it, given our positive results with High-Quality data, we expect that fine-tuning on the same might lead to higher improvements. The specific hyperparameters used for language-wise and language-family adapters can be found in Appendix-\ref{appendix-a} Table-\ref{table:adapter-ft-hyperparams}.

\begin{table}[h]
\centering 
\small
\begin{tabular}{@{}cccc@{}}
\toprule
\textbf{Lang} & \textbf{base} & \textbf{LW} & \textbf{LF} \\ \midrule
as            & 19.7          & \textbf{21.0}        & 20.6        \\
bn            & 28.8          & 28.8        & \textbf{29.2}        \\
gu            & 30.6          & \textbf{30.8}        & \textbf{30.8}        \\
hi            & 34.1          & \textbf{34.4}        & 34.2          \\
kn            & \textbf{26.1}          & \textbf{26.1}        & \textbf{26.1}        \\
ml            & \textbf{28.2}          & \textbf{28.2}        & 27.9        \\
mr            & \textbf{28.1}          & 28.0        & 27.7        \\
or            & 26.8          & 26.7        & \textbf{27.2}          \\
pa            & 31.2          & \textbf{31.3}        & 31.2        \\
ta            & \textbf{25.1}          & 25.0        & \textbf{25.1}          \\
te            & 30.4          & \textbf{30.7}        & 30.4        \\ \midrule
Avg           & 28.1          & 28.3        & 28.1        \\ \bottomrule
\end{tabular}
\caption{Results of language-wise (LW) and language-family (LF) adapter fine-tuning  of \textit{base} SLD model.}
\label{table:adapter-FT}
\end{table}

\subsection{Key Takeaways and Recommendations}
We have the following lessons:

\noindent \textbf{1.} The use of active learning techniques produced comparable results, and no single approach stood out as the best. Batch-Level distillation exhibited the strongest numerical performance, but the improvements were statistically insignificant.

\noindent \textbf{2.} Multiple metrics should be used to evaluate translations. Paraphrases of the target did not score well in BLEU but were rated highly with Chrf++.

\noindent \textbf{3.} Multistage training, involving complete dataset training followed by fine-tuning on a High-Quality fraction, improves model performance. To maintain consistent distribution, the proportions of translation pairs from each language should be similar during data filtering, and the length distribution should resemble the original dataset.

\noindent \textbf{4.} The use of adapters did not improve model performance, attributed to insufficient parameterization. With learning rate and batch size tuning, equal language family proportions should be maintained during multilingual adapter fine-tuning.

\noindent \textbf{5.} Narrower but deeper models can achieve comparable performance to wider but shallower models, despite having fewer parameters. Increasing depth by adding layers can lead to diminishing returns with increasing inference latency.

\noindent \textbf{6.} Recurrently-stacked networks, despite their promise, do not deliver in multilingual settings like ours with low to high-resource languages. However, multi-stage training is recommended for such models and, generally, for lower-parameter ones.
\section{Conclusion and Future Work}
In this paper we have empirically studied the compression of MNMT models, taking Indic to English translation as a case study, and explored the effectiveness of prominent knowledge distillation approaches. We have also studied the impact of model size, parameter sharing, multi-stage training, and quality of training data. We confirm the high difficulty of this task but make several recommendations that we expect will benefit practitioners. Having noted the positive impact of High-Quality data, we will explore this aspect in further detail in the future. We will also expand to MNMT models focusing on other language groups. Finally, the impact of post-training quantization approaches and low-precision decoding will also be investigated.
\section{Acknowledgements}
We sincerely thank Prof. Mitesh Khapra and Pranjal Agadh Chitale for their valuable insights and comments on the paper. We also extend our appreciation to the Center for Development of Advanced Computing\footnote{\url{https://www.cdac.in/index.aspx?id=print_page&print=PN}} (CDAC) for providing us with the necessary computing resources to conduct our experiments.

\bibliography{anthology, eamt23}
\bibliographystyle{eamt23}

\clearpage
\appendix

\section{Hyperparameter Details} \label{appendix-a}
\begin{table}[h]
\centering 
\small
\begin{tabular}{@{}ll@{}}
\toprule
\textbf{Hyperparameter}                   & \textbf{Value} \\ \midrule
Global Batch size                         & $64$K          \\ 
Dropout                                   & $0.2$          \\
Label smoothing                           & $0.1$          \\
Gradient clipnorm                         & $1.0$          \\
Early-stopping patience                   & $5$            \\
Optimizer                                 & Adam    \\
Adam betas                                & $(0.9, 0.98)$  \\
learning\_rate                            & $5e$-$4$       \\
lr\_scheduler                             & inverse-sqrt decay \\
Warmup steps                              & $4000$         \\ \bottomrule
\end{tabular}
\caption{Hyperparameters employed for training the student models, identical to those used for training IndicTrans}
\label{table:training-hyperparams}
\end{table}

\begin{table}[h]
\centering 
\tiny
\begin{tabular}{@{}lll@{}}
\toprule
\textbf{Hyperparameter} & \textbf{LW}                                   & \textbf{LF} \\ \midrule
Global Batch size       & $2$K (as), 8K                                 & $24$K       \\
Adapter Dropout         & $0.1$                                         & $0.1$       \\
Adapter Activation      & GELU                                          & GELU        \\
Adapter Bottleneck      & $256$                                         & $256$       \\
learning\_rate          & $1e$-$3$                                      & $1e$-$3$    \\
Warmup steps            & $1000$ (as), $2000$ (gu), $1600$ (or), $4000$ & $4000$      \\ \bottomrule
\end{tabular}
\caption{Hyperparameters employed for Adapter fine-tuning. Note that, the rest of the model hyperparameters are the same as in Table-\ref{table:training-hyperparams}}
\label{table:adapter-ft-hyperparams}
\end{table}

\section{Additional Analysis} \label{appendix-b}
This section presents the remaining Chrf++ results for Distillation techniques, Adapter fine-tuning, Width-vs-Height Analysis, and Multistage training.

\begin{table}[h]
\centering 
\small
\setlength{\tabcolsep}{2pt}
\begin{tabular}{@{}ccc|ccccc@{}}
\toprule
\textbf{Lang} & \textbf{OG\_base} & \textbf{IT} & \textbf{SLD} & \textbf{W+S LD} & \textbf{BL}   & \textbf{GL} & \textbf{GLwD} \\ \midrule
as            & 43.0              & 48.2        & 44.8          & 44.9            & \textbf{45.5} & 45.2        & 45.1          \\
bn            & 54.6              & 56.9        & 54.7          & 54.6            & \textbf{55.0} & 54.3        & 54.6          \\
gu            & 55.9              & 58.7        & 56.2          & 56.8            & \textbf{56.9} & 56.6        & 56.5          \\
hi            & 58.9              & 61.3        & 58.7          & 59.0            & \textbf{59.3} & 59.0        & 59.0          \\
kn            & 51.4              & 54.6        & \textbf{52.2} & 52.1            & \textbf{52.2} & 52.1        & \textbf{52.2} \\
ml            & 53.6              & 57.2        & 54.3          & 54.3            & \textbf{54.6} & 53.9        & 54.4          \\
mr            & 53.2              & 56.4        & 54.0          & 53.9            & \textbf{54.2} & 53.7        & 53.6          \\
or            & 52.2              & 55.5        & 53.0          & \textbf{53.2}   & 52.9          & 53          & 52.8          \\
pa            & 56.2              & 60.0        & 56.4          & 56.7            & \textbf{56.9} & 56.8        & 56.7          \\
ta            & 51.1              & 54.1        & 51.1          & 51.1            & \textbf{51.3} & 51.2        & \textbf{51.3} \\
te            & 55.3              & 58.2        & 55.7          & \textbf{55.9}   & 55.7          & 55.8        & 55.8          \\ \midrule
Avg           & 53.2              & 56.5        & 53.7          & 53.9            & 54.0          & 53.8        & 53.8          \\ \bottomrule
\end{tabular}
\caption{Chrf++ scores of base model \textit{distilled} with various distillation techniques. Note that the IndicTrans (IT) scores in the first column are for reference.}
\label{table:distillation-chrf-results}
\end{table}

\begin{table}[h]
\centering 
\small
\begin{tabular}{@{}cccc@{}}
\toprule
\textbf{Lang} & \textbf{base} & \textbf{LW} & \textbf{LF} \\ \midrule
as            & \textbf{45.8}          & 45.6        & 45.1        \\
bn            & 54.7          & 54.7        & \textbf{54.9}        \\
gu            & 56.2          & \textbf{56.4}        & 56.3        \\
hi            & 58.7          & \textbf{58.8}        & 58.7        \\
kn            & 52.2          & \textbf{52.4}        & 52.2        \\
ml            & \textbf{54.3}          & 54.2        & 54.1        \\
mr            & \textbf{54.0}          & 53.8        & 53.7        \\
or            & \textbf{53.0}          & 52.7        & \textbf{53.0}        \\
pa            & \textbf{56.4}          & 56.3        & 56.2        \\
ta            & \textbf{51.1}          & 50.9        & 50.8          \\
te            & 55.7          & \textbf{55.9}        & 55.6        \\ \midrule
Avg           & 53.7          & 53.8        & 53.7        \\ \bottomrule
\end{tabular}
\caption{Chrf++ Results of language-wise (LW) and language-family (LF) adapter fine-tuning  of \textit{base} SLD model.}
\end{table}

\begin{table}[h]
\centering 
\small
\begin{tabular}{@{}ccccc@{}}
\toprule
\textbf{Lang} & \textbf{huge$_{RS}$} & \textbf{base$12L$} & \textbf{base$18L$} & \textbf{base$24L$} \\ \midrule
as                 & 42.9                   & 46.6               & \textbf{48.0}                 & 47.9               \\
bn                 & 52.9                   & 55.4               & 56.3               & \textbf{56.4}               \\
gu                 & 55.2                   & 58.0               & 58.6               & \textbf{58.8}               \\
hi                 & 58.4                   & 60.1               & \textbf{60.5}      & 60.3                        \\
kn                 & 51.2                   & 53.2               & \textbf{54.1}      & \textbf{54.1}               \\
ml                 & 52.5                   & 55.4               & 55.8               & \textbf{56.3}               \\
mr                 & 52.0                   & 55.1               & 55.9               & \textbf{56.2}               \\
or                 & 50.7                   & 54.3               & 55.3               & \textbf{55.5}               \\
pa                 & 56.1                   & 58.1               & 58.7               & \textbf{59.0}                 \\
ta                 & 50.1                   & 52.3               & 53.1               & \textbf{53.5}               \\
te                 & 54.2                   & 56.6               & 57.7               & \textbf{57.9}               \\ \midrule
Avg                & 52.4                  & 55.0              & 55.8              & 56.0              \\ \bottomrule
\end{tabular}
\caption{Chrf++ scores for Width-vs-Height analysis}
\end{table}

\begin{table}[h]
\centering
\small
\begingroup
\setlength{\tabcolsep}{1.25pt} 
\begin{tabular}{@{}ccccccc@{}}
\toprule
\textbf{Lang} & \textbf{base} & \textbf{base$12L$} & \textbf{base$18L$} & \textbf{base$24L$} & \textbf{big} & \textbf{huge$_{RS}$} \\ \midrule
as & 20.3 & 22.3 & \textbf{23.6} & 23.2 & 23.3 & 20.4 \\
bn & 29.0 & 30.3 & 31.2 & \textbf{31.6} & 31.1 & 28.6 \\
gu & 31.2 & 33.1 & 34.0 & 34.1 & \textbf{34.2} & 31.5 \\
hi & 34.3 & 36.1 & \textbf{36.8} & 36.6 & 36.5 & 35.1 \\
kn & 26.4 & 27.6 & \textbf{28.5} & \textbf{28.5} & 28.1 & 26.2 \\
ml & 28.7 & 29.9 & 30.6 & \textbf{31.1} & 30.6 & 28.1 \\
mr & 28.1 & 30.0 & 30.8 & 30.9 & \textbf{31.2} & 27.9 \\
or & 27.3 & 28.9 & 29.3 & 29.9 & \textbf{30.1} & 26.7 \\
pa & 31.5 & 33.3 & 34.4 & \textbf{34.8} & 34.3 & 32.2 \\
ta & 25.3 & 26.9 & 27.5 & \textbf{28.1} & 27.7 & 25.4 \\
te & 30.6 & 31.8 & \textbf{33.5} & \textbf{33.5} & 33.3 & 30.2 \\ \midrule
Avg & 28.4 & 30.0 & 30.9 & 31.1 & 30.9 & 28.4 \\ \bottomrule
\end{tabular}
\endgroup
\caption{Absolute BLEU scores obtained by Multi-stage training.}
\label{hqft-abs-bleu}
\end{table}

\begin{table}[h]
\centering
\tiny
\begin{tabular}{@{}ccccccc@{}}
\toprule
\textbf{Lang} & \textbf{base} & \textbf{base$12L$} & \textbf{base$18L$} & \textbf{base$24L$} & \textbf{big} & \textbf{huge$_{RS}$} \\ \midrule
as            & 45.5 (0.7)    & 47.5 (0.9)         & 48.7 (0.7)         & 48.5 (0.6)         & 48.2 (0.1)   & 44.3 (1.4)           \\
bn            & 55.0 (0.3)    & 55.9 (0.5)       & 56.6 (0.3)           & 56.8 (0.4)         & 56.6 (0.2)   & 54.1 (1.2)           \\
gu            & 56.9 (0.7)    & 58.4 (0.4)         & 59.0 (0.4)         & 59.1 (0.3)         & 58.9 (0.5)   & 56.5 (1.3)           \\
hi            & 59.1 (0.4)    & 60.2 (0.1)         & 60.8 (0.3)         & 60.7 (0.4)         & 60.8 (0.4)   & 59.4 (1.0)             \\
kn            & 52.5 (0.3)    & 53.7 (0.5)         & 54.5 (0.4)         & 54.7 (0.6)         & 54.1 (0.3)   & 52.2 (1.0)             \\
ml            & 54.9 (0.6)    & 56.1 (0.7)         & 56.6 (0.8)         & 57.0 (0.7)           & 56.8 (0.7)   & 54.0 (1.5)             \\
mr            & 54.3 (0.3)    & 55.9 (0.8)         & 56.4 (0.5)         & 56.6 (0.4)         & 56.7 (0.5)   & 53.6 (1.6)           \\
or            & 53.4 (0.4)    & 55.0 (0.7)         & 55.5 (0.2)         & 55.9 (0.4)         & 55.8 (0.9)   & 52.6 (1.9)           \\
pa            & 56.9 (0.5)    & 58.3 (0.2)         & 59.2 (0.5)         & 59.6 (0.6)         & 59.2 (0.3)   & 57.2 (1.1)           \\
ta            & 51.4 (0.3)    & 52.8 (0.5)         & 53.3 (0.2)         & 54.0 (0.5)           & 53.6 (0.4)   & 51.2 (1.1)           \\
te            & 56.1 (0.4)    & 57.2 (0.6)         & 58.1 (0.4)         & 58.4 (0.5)         & 58.1 (0.5)   & 55.2 (1.0)             \\ \midrule
Avg           & 54.1 (0.5)  & 55.5 (0.5)       & 56.2 (0.4)       & 56.5 (0.5)       & 56.2 (0.4) & 53.7 (1.3)         \\ \bottomrule
\end{tabular}
\caption{Multistage training Chrf++ results. The bracketed number denotes the Chrf++ improvement due to High-Quality fine-tuning.}
\end{table}
\clearpage
\section{Note on Evaluation} \label{appendix-c}
This paper mainly relies on BLEU and Chrf++, but lately, COMET\footnote{\url{https://unbabel.github.io/COMET/html/index.html}} is becoming popular. However, COMET is unavailable for most Indic languages we study. Therefore, we leave this for future work.
\end{document}